\documentclass[11pt,a4paper]{article}
\usepackage[utf8]{inputenc} 
\usepackage[T1]{fontenc}    
\usepackage[hyperref]{emnlp2020}
\usepackage{times}
\usepackage{latexsym}
\usepackage{amsmath,amssymb,amsthm}
\usepackage{amsfonts}       
\usepackage{mathtools}
\usepackage{graphicx}
\usepackage{relsize}
\usepackage{booktabs}
\usepackage{textcomp}

\usepackage{microtype}

\aclfinalcopy 

\usepackage{amsmath,amsfonts,bm}

\newcommand{\secref}[1]{\S\ref{#1}}
\def\Secref#1{Section~\ref{#1}}

\def\eqref#1{equation~\ref{#1}}

\def\1{\bm{1}}

\def\vtheta{{\bm{\theta}}}

\def\vc{{\bm{c}}}
\def\vd{{\bm{d}}}
\def\ve{{\bm{e}}}

\def\vg{{\bm{g}}}

\def\vo{{\bm{o}}}

\DeclareMathAlphabet{\mathsfit}{\encodingdefault}{\sfdefault}{m}{sl}
\SetMathAlphabet{\mathsfit}{bold}{\encodingdefault}{\sfdefault}{bx}{n}

\newcommand{\topdata}{\texttt{TOP}}
\newcommand{\mdtop}{\texttt{TOPv2}}

\newcommand{\decoupled}{\textsc{S{\smaller EQ}2S{\smaller EQ}-C{\smaller OPY}P{\smaller TR}}}
\newcommand{\copyptr}{C{\smaller OPY}P{\smaller TR}}
\newcommand{\domS}{\mathbb{S}} 

\newcounter{magicrownumbers}
\newcommand{\rownumber}[1]{\refstepcounter{magicrownumbers}\arabic{magicrownumbers}\label{#1}}
\newcommand{\ftextnumero}{{\fontfamily{txr}\selectfont \textnumero}}

\title{Low-Resource Domain Adaptation for\\Compositional Task-Oriented Semantic Parsing}

\author{
Xilun Chen \And Asish Ghoshal \AND
Yashar Mehdad \And Luke Zettlemoyer \And Sonal Gupta \AND
\textnormal{Facebook Inc.}\\
\texttt{\{xilun,aghoshal,mehdad,lsz,sonalgupta\}@fb.com}
}

\date{}

\begin{document}
\maketitle
\begin{abstract}

Task-oriented semantic parsing is a critical component of virtual assistants, which is responsible for understanding the user's intents (set reminder, play music, etc.).
Recent advances in deep learning have enabled several approaches to successfully parse more complex queries~\cite{gupta-etal-2018-semantic-parsing,rongali-etal-2020-dont}, but these models require a large amount of annotated training data to parse queries on new \emph{domains} (e.g.~reminder, music).

In this paper, we focus on adapting task-oriented semantic parsers to low-resource domains, and propose a novel method that outperforms a supervised neural model at a 10-fold data reduction.
In particular, we identify two fundamental factors for low-resource domain adaptation: better \emph{representation learning} and better \emph{training techniques}.
Our representation learning uses BART~\cite{lewis2019bart} to initialize our model which outperforms encoder-only pre-trained representations used in previous work.
Furthermore, we train with optimization-based meta-learning~\cite{finn-etal-2017-maml} to improve generalization to low-resource domains.
This approach significantly outperforms all baseline methods in the experiments on a newly collected multi-domain task-oriented semantic parsing dataset (\mdtop{}\footnote{The dataset can be downloaded at \url{https://fb.me/TOPv2Dataset}}).

\end{abstract}


\section{Introduction}\label{sec:intro}

Virtual Assistants now play an ever increasingly important role in our daily life, and can help users perform a wide spectrum of tasks ranging from setting personal reminders, checking local weather, to controlling smart home devices and online shopping.
A critical step in any virtual assistant is to understand the user's intent (e.g.\ set reminder, get weather info, etc.) given the user utterance.
In recent years, a number of successful models have emerged to tackle such task-oriented semantic parsing task, for both simple and more complex queries.

\begin{figure}
    \centering
    \includegraphics[width=\linewidth]{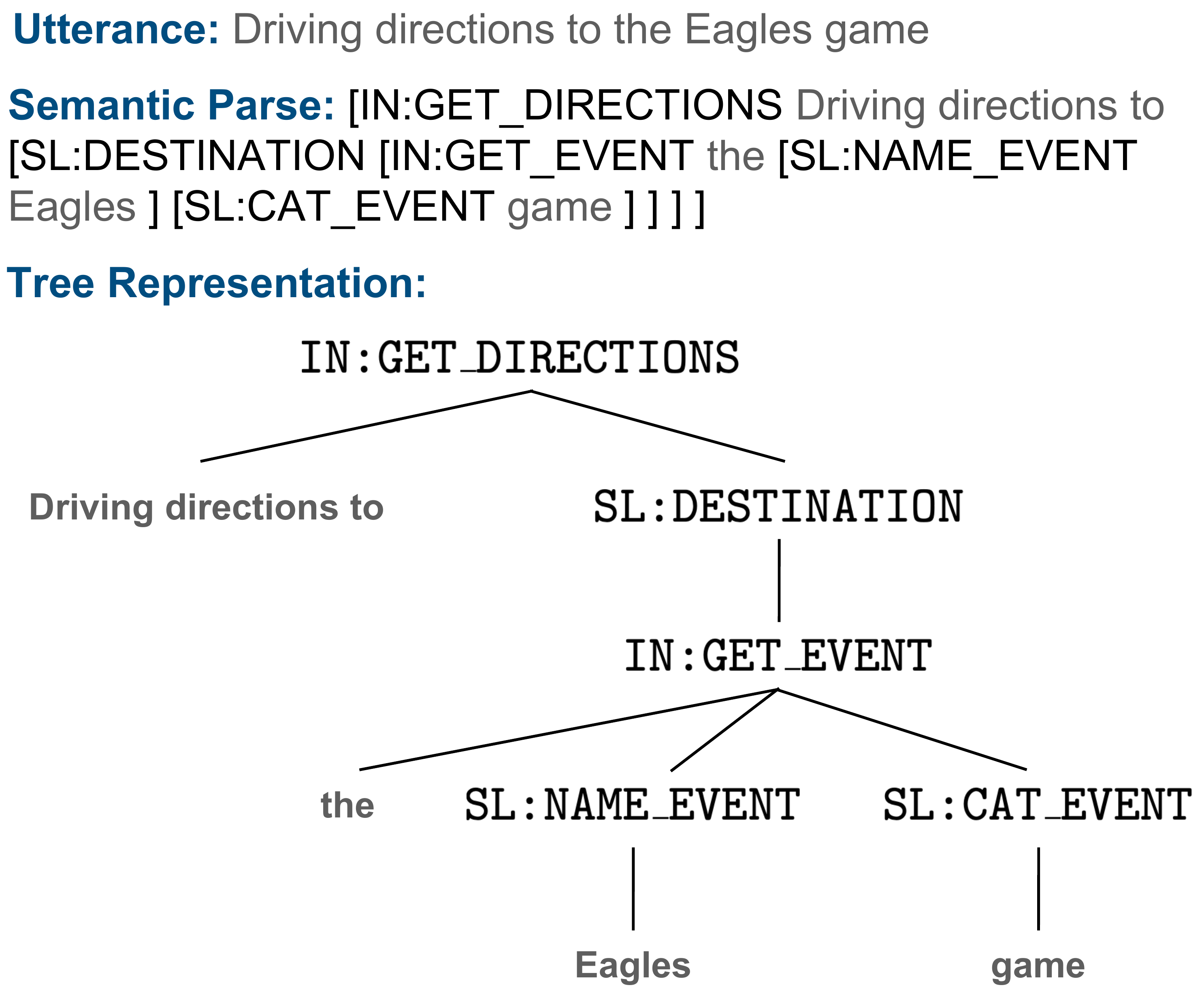}
    \caption{An compositional query from \texttt{TOP} dataset.}
    \label{fig:nested-query}
\end{figure}

Traditionally, task-oriented semantic parsers treat the problem as a joint \emph{intent classification} and \emph{slot filling} task~\cite{Liu+2016}, where the model first predicts the intent of the input utterance from a set of pre-defined intent labels, and then identify all the necessary slots for that intent.
For instance, for the query \emph{``How's the weather in San Francisco?''}, the model would predict the \texttt{GET\_WEATHER} intent, and tag \emph{San Francisco} as a \texttt{LOCATION} slot.
With the elevated expectation of virtual assistants, however, techniques for handling the more complex \emph{compositional} queries have been proposed recently using neural parsers~\cite{gupta-etal-2018-semantic-parsing} or Seq2Seq models~\cite{rongali-etal-2020-dont}.
In particular, these approaches can handle complicated queries with multiple intents or nested slots.
For example, the following query from the \texttt{TOP} dataset~\cite{gupta-etal-2018-semantic-parsing} ``\emph{Driving directions to the Eagles game}'' is composed of a \texttt{GET\_DIRECTIONS} intent and a \texttt{GET\_EVENT} one nested in a tree structure (Figure~\ref{fig:nested-query}).

On the other hand, most of these deep neural models require a large amount of annotated training data to achieve a good performance, which is aggravated by the fact that virtual assistants need to support hundreds of tasks each mandating a separate set of labeled training samples.
Therefore, in order to support more diverse use cases without an excessive need in human annotated data, it becomes crucial that the semantic parsing model has the capability to generalize to new \emph{tasks} or \emph{domains} (reminder, music, etc.) with a limited number of labeled samples in the target domain.
While transfer learning methods have been proposed for traditional sequence tagging models~\cite{Jaech2016DomainAO,goyal-etal-2018-fast} to help facilitate learning slot filling models for domains with less annotated data, the efforts have been lacking when it comes to developing compositional semantic parsers in such low-resource domain adaptation setting.

Therefore, we in this paper show it is possible to build compositional task-oriented semantic parsers for low-resource domains (e.g.\ 25 training samples per intent or slot label), and propose a solution that is competitive against supervised models trained with 10x more data.
We identify two key factors for successfully adapting task-oriented semantic parsers to new domains: better \textbf{representation learning} and better \textbf{training techniques}.

We first show that pre-trained language representations are critical in the low-resource setting for the model to quickly generalize to new intents and slots.
Furthermore, most pre-trained language representations used in previous work such as BERT~\cite{devlin-etal-2019-bert} or RoBERTa~\cite{DBLP:journals/corr/abs-1907-11692} are encoder-only models, and are hence not ideal for a compositional parser with an encoder-decoder (seq2seq) architecture.
We therefore propose to use BART~\cite{lewis2019bart}, a pre-trained seq2seq model that can be used to initialize both the encoder and decoder of our semantic parser, which significantly outperforms other pre-trained representations such as RoBERTa.

More importantly, these large pre-trained models are sometimes known to pose challenges to fine-tuning with very few training samples.
In order to better adapt the semantic parser to low-resource domains, we employ \emph{optimization-based meta-learning}~\cite{finn-etal-2017-maml} to improve generalization of the BART model trained on the source domains, making it easier to be fine-tuned on the target domains with very little training data.

Finally, in order to evaluate our approach, we collect a multi-domain compositional task-oriented semantic parsing dataset (\mdtop{}), based on the original \texttt{TOP}~\cite{gupta-etal-2018-semantic-parsing} dataset.
In addition to the \emph{navigation} and \emph{event} domains found in \texttt{TOP}, our \mdtop{} dataset has 6 new domains: \emph{alarm, messaging, music, reminder, timer}, and \emph{weather}, with more than $137k$ new samples.
We conduct extensive experiments on this new dataset, showing that our proposed method significantly outperforms all the baseline models in the low data regime.
We further show that our model achieves competitive performance compared to supervised state-of-the-art models while using 10x less data.
\section{Problem Setup}\label{sec:model:setup}

\begin{figure*}
    \centering
    \includegraphics[width=\linewidth]{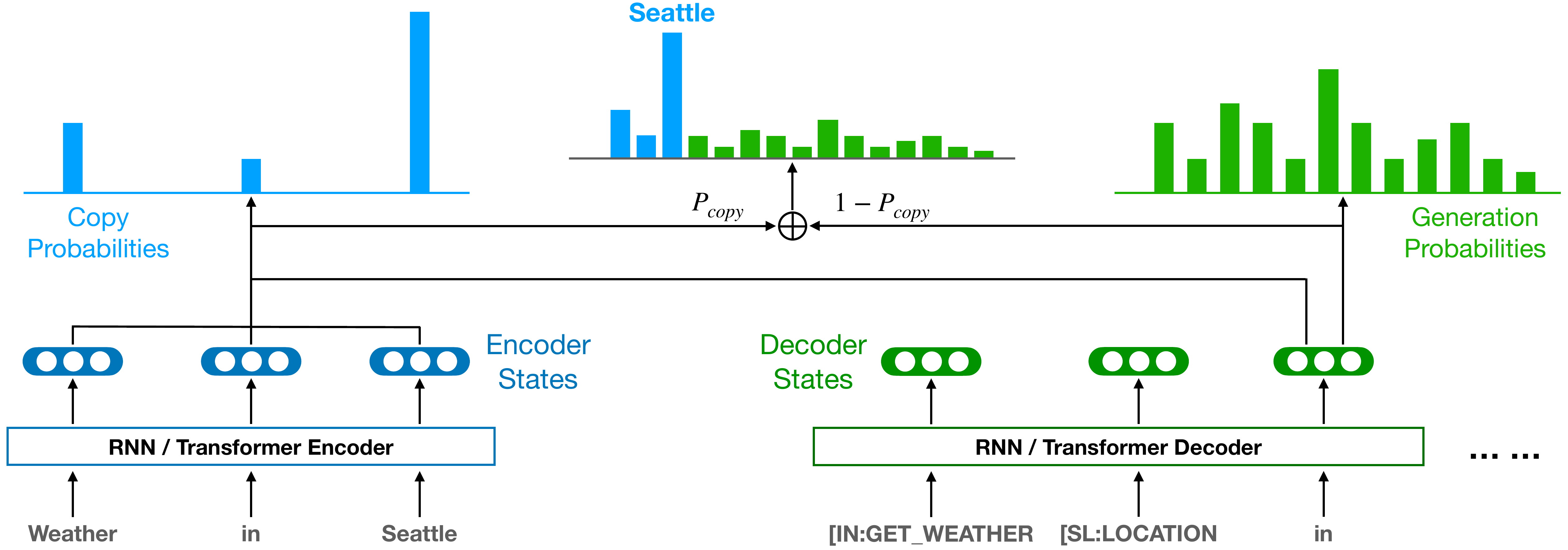}
    \caption{Overview of our sequence-to-sequence architecture with copy pointer (\decoupled). Note that the target sequence is slightly different from original \topdata{} dataset, as explained in the pre-processing steps in \secref{sec:exp:setup}.}
    \label{fig:model-arch}
\end{figure*}

We first formally define the task of domain adaptation (or \emph{domain scaling}) for task-oriented semantic parsing.
As illustrated in Figure~\ref{fig:nested-query}, the task-oriented parsing task aims to predict the semantic parse given the \emph{user utterance} (or \emph{query}).
The \emph{semantic parse} has a tree structure and is represented using a serialized tree representation defined in the \topdata{} dataset~\cite{gupta-etal-2018-semantic-parsing}.
Following recent state-of-the-art practices~\cite{rongali-etal-2020-dont}, we formulate the problem as a sequence-to-sequence (seq2seq) task, where the utterance is treated as the \emph{source sequence} $\cal S$, while the semantic parse serves as the \emph{target sequence} $\cal T$.

In our domain scaling setting, the goal is to develop a semantic parser with minimal training examples on a set of new \emph{target domains}.\footnote{Zero-shot transfer is very challenging since a new domain has unique otherwise unseen intents and slots. Previous work~\cite{DBLP:conf/aaai/LeeJ19} on zero-shot transfer relied on additional prior knowledge such as slot descriptions.}
Formally, denote $\mathbb{T} = \{ D_1^T, ..., D_N^T\}$ as the set of $N$ target domains.
On the other hand, we assume access to training data for a number of \emph{source domains}, which can be used to help build models on the target domains.
Denote $\mathbb{S} = \{ D_1^S, ..., D_M^S\}$ as the set of $M$ source domains.
For each source domain $d\in \mathbb{S}$, there exists a set of annotated training data $D^d=(\mathcal{S}_d, \mathcal{T}_d)$ where $\mathcal{S}_d$ and $\mathcal{T}_d$ are the utterance and semantic parse, respectively.

For a target domain $t\in \mathbb{T}$, however, only a very limited number of training instances exist.
As domains differ drastically in terms of complexity (shown in Table~\ref{tab:mdtop}), we would expect models to require varying amount of training data for each.
To normalize for such effects, we introduce a new task-specific measure of training set size: \textbf{SPIS}, which stands for \emph{samples per intent and slot}, indicating the number of training samples available for each intent and slot label.
Traditionally, at least a few hundreds of training samples are needed for each label to successfully train a deep neural semantic parser (see \secref{sec:exp:result}).
In the low-resource setting, in contrast, we focus on a scenario with much less training data at 25 SPIS.
That is, for each intent and slot in the target domain, 25 training samples are required\footnote{Note that one sample may contain multiple intents and slots. Empirically, only around 10 distinct samples are selected for each intent and slot (Table~\ref{tab:mdtop-results}).}.
(See \secref{sec:exp:spis-curve} for discussions on more SPIS settings.)
To benchmark the performance on the low-resource domains, we compare with various high-resource supervised baselines trained with much more data up to 500 or 1000 SPIS.

\section{Base Model}\label{sec:model:arch}

In this section, we present our core model architecture.
Our meta-learning technique will be introduced in \Secref{sec:metalearning}.
We follow recent state-of-the-art approaches~\cite{rongali-etal-2020-dont,session-based-decoupled} and adopt a seq2seq model as our base architecture (\decoupled), derived from the Pointer Generator Network~\cite{see-etal-2017-get}.
The base architecture is shown in Figure~\ref{fig:model-arch}.

For an input sequence $\mathcal{S}=[ w_1, w_2, \dots, w_n]$, the encoder first encodes it into a series of hidden vectors (encoder states) $[ \ve_1, \ve_2, \dots, \ve_n]$.
The encoder states are then passed to an decoder that autoregressively produces target tokens $\vo_t$ for each timestamp $t$.
Specifically, the decoder first outputs a hidden decoder state $\vd_t$ based on the decoder states from previous timestamps as well as all encoder states:
\begin{equation*}
    \vd_t = Decoder(\ve_1,\dots,\ve_n ; \vd_1,\dots,\vd_{t-1})
\end{equation*}
In task-oriented semantic parsing, the target sequence consists of two types of tokens: \emph{utterance tokens} that always come from the source sequence, and \emph{ontology tokens} that represent intent and slot labels.
Therefore, two probability distributions are formulated and combined in order to produce the output token, namely the \emph{copy probability} and the \emph{generation probability}.
The generation probability $\vg_t$ is produced by the decoder by mapping the decoder state onto the output vocabulary, which only includes ontology tokens but not utterance tokens.
\begin{equation*}
    \vg_t = softmax(OutputEmbed(\vd_t))
\end{equation*}
The copy probability $\vc_t$, on the other hand, indicates whether to copy one of the source tokens as the decoder output for timestamp $t$, and is predicted by using $\vd_t$ as the query to perform a multi-head attention \citep[MHA,][]{NIPS2017_7181} over the encoder states.
\begin{align*}
    \vc_t, {\bm \omega}_t &= \mathrm{MHA}(\ve_1,\dots,\ve_n, Linear(\vd_t))\\
    P_{copy} &= sigmoid(Linear([\vd_t; {\bm\omega}_t]))
\end{align*}
where $\vc_t$ are the attention weights indicating the copy probability and ${\bm\omega}_t$ is the attended vector used to compute a scalar $P_{copy}$ to weigh between copying and generation when constructing the final output token $\vo_t$:
\begin{equation*}
    \vo_t = P_{copy} \cdot \vc_t + (1-P_{copy}) \cdot \vg_t
\end{equation*}

\subsection{Pre-trained Language Representations}\label{sec:model:langrepr}
In \secref{sec:model:arch}, we introduce a general model framework where the encoder and decoder can be any RNN or Transformer~\cite{NIPS2017_7181} architecture.
In practice, pre-trained language representations such as BERT~\cite{devlin-etal-2019-bert} have greatly improved performance across many NLP tasks.
For task-oriented semantic parsing, in particular, \newcite{rongali-etal-2020-dont} achieved state-of-the-art performance with RoBERTa~\cite{DBLP:journals/corr/abs-1907-11692}.
Departing from previous work, we argue that such encoder-only pre-trained models are not the most suitable choice for a seq2seq semantic parser, as it couples a pre-trained encoder with a randomly initialized decoder, resulting in challenges during training.
Instead, we adopt BART~\cite{lewis2019bart}, a pre-trained seq2seq model, which can be used to initialize both the encoder and decoder in our \decoupled{} model.

\subsection{Training Stages}\label{sec:model:stages}
Finally, we clarify the terminology adopted for various training stages.
Several training strategies exist for domain adaptation.
For instance, one can employ \emph{joint training} that trains a single model with all the available data on both source and target domains (with optional upsampling on the target domains).
Another approach, which we found superior (Table~\ref{tab:mdtop-results}), is the \emph{pre-training + fine-tuning} strategy, where a model is first trained on the source domains and then fine-tuned on the low-resource target domains.

On the other hand, as pre-trained language representations such as RoBERTa or BART are adopted, the latter strategy becomes a 3-stage training process: train RoBERTa/BART; fine-tune on the source domains; fine-tune again on the target domains.
To avoid confusion, we standardize the terminology used to refer to each of these three stages.
The first stage, which is out of scope for this paper, is the \textbf{pre-training} stage where self-supervised language representations are learned.
Then, it is fine-tuned on the source domains. We call this stage \textbf{source training} to avoid ambiguity with the final stage.
In the final stage, which is denoted as \textbf{fine-tuning}, the source-trained model is fine-tuned again on the target domains.
It is possible to omit the second stage and directly fine-tune pre-trained RoBERTa/BART on the target domains.
As shown in Table~\ref{tab:mdtop-results}, however, source-training significantly improves the final performance.

\section{Meta Learning}\label{sec:metalearning}

As mentioned in \secref{sec:model:stages}, model training for low-resource domain adaptation consists of two stages: source training and fine-tuning (or target training), where the model (initialized with pre-trained representations) is first trained on the source domains and then fine-tuned on each low-resource target domain.
As target domain training data is scarce, it might be challenging to effectively fine-tune a large BART model with only 25 samples per intent and slot.
One reason is that traditional source training optimizes the model performance solely for the source domains, which may result in a model with strong performance on the source domains but less than ideal for transferring to new target domains via fine-tuning.

\begin{table*}
\centering
\small
\begin{tabular}{@{\hspace{0.4em}}l@{\hspace{0.4em}}@{\hspace{0.4em}}r@{\hspace{0.4em}}@{\hspace{0.4em}}r@{\hspace{0.4em}}@{\hspace{0.4em}}r@{\hspace{0.4em}}@{\hspace{0.4em}}r@{\hspace{0.4em}}@{\hspace{0.4em}}r@{\hspace{0.4em}}@{\hspace{0.4em}}r@{\hspace{0.4em}}@{\hspace{0.4em}}r@{\hspace{0.4em}}@{\hspace{0.4em}}l@{\hspace{0.4em}}}
\toprule
Domain  & \#Train    & \#Valid    & \#Test    & \#Int &   \#Slt &   Flat\%  &   Depth & Example Utterance \\
\midrule
alarm       & 20430 & 2935  & 7123  & 8     & 9     & 84\%  & 2.16  & \emph{Set alarm for noon tomorrow.} \\
event       & 9170  & 1336  & 2654  & 11    & 17    & 80\%  & 2.37  & \emph{whats happening in san francisco tonight} \\
messaging   & 10018  & 1536  & 3048  & 12     & 27    & 84\%  & 2.23  & \emph{Text yes to Bill and Mindy.} \\
music       & 11563 & 1573  & 4184  & 15    & 9     & 100\% & 1.98  & \emph{Repeat the last album} \\
navigation  & 20998 & 2971  & 6075  & 17    & 33    & 57\%  & 2.68  & \emph{I need to know if there's a lot of traffic on my way home} \\
reminder    & 17840 & 2526  & 5767  & 19    & 32    & 79\%  & 2.45  & \emph{erase reminder to attend conference this monday} \\
timer       & 11524 & 1616  & 4252  & 11    & 5     & 96\%  & 2.00  & \emph{Set a timer for 2 hours} \\
weather     & 23054 & 2667  & 5682  & 7     & 11    & 100\% & 1.93  & \emph{how cold is it?} \\[0.4em]
\bf total   & 125k  & 17k & 39k & 80 & 82 & 84\% & 2.24 &\\

\specialrule{\heavyrulewidth}{\aboverulesep}{\belowrulesep}
Domain      & \multicolumn{8}{l}{\hspace{-2mm}\textbf{Canonical} semantic parse for the example utterance (see \secref{sec:exp:setup})} \\
\midrule
alarm       & \multicolumn{8}{l}{\hspace{-2mm}\texttt{[IN:CREATE\_ALARM} \texttt{[SL:DATE\_TIME} for noon tomorrow \texttt{]} \texttt{]}} \\
event       & \multicolumn{8}{l}{\hspace{-2mm}\texttt{[IN:GET\_EVENT} \texttt{[SL:DATE\_TIME} tonight \texttt{]} \texttt{[SL:LOCATION} san francisco \texttt{]} \texttt{]}} \\
messaging   & \multicolumn{8}{l}{\hspace{-2mm}\texttt{[IN:SEND\_MESSAGE} \texttt{[SL:CONTENT\_EXACT} yes \texttt{]} \texttt{[SL:RECIPIENT} bill \texttt{]} \texttt{[SL:RECIPIENT} mindy \texttt{]} \texttt{]}}\\
music       & \multicolumn{8}{l}{\hspace{-2mm}\texttt{[IN:REPLAY\_MUSIC} \texttt{[SL:MUSIC\_TYPE} album \texttt{]} \texttt{]}}\\
navigation  & \multicolumn{8}{l}{\hspace{-2mm}\texttt{[IN:GET\_INFO\_TRAFFIC} \texttt{[SL:DESTINATION} \texttt{[IN:GET\_LOCATION\_HOME} \texttt{]} \texttt{]} \texttt{]}} \\
reminder    & \multicolumn{8}{l}{\hspace{-2mm}\texttt{[IN:DELETE\_REMINDER} \texttt{[SL:DATE\_TIME} this monday \texttt{]} \texttt{[SL:TODO} attend conference \texttt{]} \texttt{]}} \\
timer       & \multicolumn{8}{l}{\hspace{-2mm}\texttt{[IN:CREATE\_TIMER} \texttt{[SL:DATE\_TIME} for 2 hours \texttt{]} \texttt{[SL:METHOD\_TIMER} timer \texttt{]} \texttt{]}} \\
weather     & \multicolumn{8}{l}{\hspace{-2mm}\texttt{[IN:GET\_WEATHER} \texttt{[SL:WEATHER\_ATTRIBUTE} cold \texttt{]} \texttt{]}} \\

\bottomrule
\end{tabular}
\caption{Statistics of the \mdtop{} dataset. \#Int: \emph{number of intents}; \#Slt: \emph{number of slots}; Flat\%: \emph{percentage of flat ($depth\leq 2$) queries}; Depth: \emph{average depth of queries}. (The example in Figure~\ref{fig:nested-query} has a depth of 4.)}
\label{tab:mdtop}
\end{table*}

Therefore, we propose to replace source training with optimization-based meta-learning~\cite{finn-etal-2017-maml} to improve generalization.
Instead of directly optimizing towards source domain accuracy, meta-learning, when trained on the source domains, looks for a good \emph{initialization} $\vtheta_0$ that can easily be adapted to new tasks (domains) with minimal fine-tuning.
In order to learn a model that is easier to be fine-tuned on new tasks with a small amount of training data, meta-learning adopts a different training objective that explicitly optimizes for generalization by \emph{repeatedly simulating low-resource fine-tuning} during training.

Specifically, in each iteration (episode), the MAML~\cite{finn-etal-2017-maml} algorithm samples two batches of training samples from a source domain $d\in\domS$: $D_s^d$ and $D_q^d$, conventionally named the \emph{support} and \emph{query} set respectively.
In standard source training, one simply computes the loss on $D_s^d$ and takes a gradient step to update the model.
In MAML, however, low-resource fine-tuning is simulated at each training episode.
Let $\vtheta$ denote the model parameters being meta-learned, MAML first takes a gradient step on $D_s^d$ that leads to:
\begin{equation*}
    \vtheta^d \gets \vtheta - \eta \nabla_{\vtheta} \mathcal{L}(\vtheta; D_s^d)
\end{equation*}
where $\eta$ is the \emph{inner} learning rate.
$\vtheta^d$ can be viewed as a minimally fine-tuned model with only one fine-tuning iteration on the source domain $d$.
We then use $D_q^d$ to evaluate how well $\vtheta^d$ generalizes to new unseen data and update of our original model $\vtheta$ with this generalization loss:
\begin{align}
    \label{eqn:maml-update}
    \vtheta &\gets \vtheta - \alpha \nabla_{\vtheta} \mathcal{L}(\vtheta^d; D_q^d) \\
    &= \vtheta - \alpha \nabla_\vtheta \mathcal{L}(\vtheta - \eta \nabla_{\vtheta} \mathcal{L}(\vtheta; D_s^d); D_q^d) \nonumber
\end{align}
where $\alpha$ is the \emph{outer} learning rate.
Such updates are performed repeatedly on all source domains to simulate the low-resource fine-tuning scenario, which eventually learns a better initialization that only requires a small amount of data for fine-tuning to achieve good performance on the target domains.
Also note that one can accumulate gradients from multiple episodes (domains) before updating the model $\vtheta$, but we choose to update $\vtheta$ after every episode following~\newcite{antoniou2018how}.

Finally, MAML requires the computation of second derivatives when unrolling Equation (\ref{eqn:maml-update}), which consumes too much memory for large models such as BART.
Therefore, we instead adopt a first-order meta-learning algorithm, Reptile~\cite{DBLP:journals/corr/abs-1803-02999}, which has shown comparative or even superior performance than MAML despite its simplicity~\cite{dou-etal-2019-investigating}.
In Reptile, $k>1$ batches of training instances $D_1^d,\dots,D_k^d$ are sampled for a source domain $d\in\domS$ in each episode, and the model is updated as follows:
\begin{align*}
    \vtheta^d &\gets \mathrm{Adam}^k(\vtheta; D_{1..k}^d, \eta), \\
    \vtheta &\gets \vtheta + \alpha (\vtheta^d - \vtheta),
\end{align*}
where $\mathrm{Adam}^k(. ; D^d_{1..k}, \eta)$ denotes performing $k$ consecutive updates on $D_1^d,\dots,D_k^d$ using Adam~\cite{kingma2014adam} with inner learning rate $\eta$.
Note that this surprisingly simple algorithm becomes equivalent to standard source training when $k=1$.
When $k>1$, however, Reptile behaves differently and performs similar updates compared to MAML as shown by~\newcite{DBLP:journals/corr/abs-1803-02999} using Taylor Series analysis.

\section{Experiments}\label{sec:exp}

In this section, we first introduce \mdtop{}, a multi-domain task-oriented semantic parsing dataset we are releasing to the community.
It is an extension to the \topdata{} dataset with 6 additional domains and $137k$ new samples.
We then outline the setup of our low-resource domain scaling experiments in \secref{sec:exp:setup}, and present the experimental results in \secref{sec:exp:result}.

\subsection{The \mdtop{} Dataset}\label{sec:exp:dataset}

While multiple datasets exist for task-oriented semantic parsing such as ATIS~\cite{price-1990-evaluation} or SNIPS~\cite{DBLP:journals/corr/abs-1805-10190}, the \topdata{} dataset~\cite{gupta-etal-2018-semantic-parsing} is unique in that it contains compositional queries with complex and hierarchical structures (Figure~\ref{fig:nested-query}).
On the other hand, the queries from the \topdata{} dataset are limited to only two domains, namely \emph{navigation} and \emph{event}, making it unsuited for domain scaling experiments.
To this end, we extend the \topdata{} dataset with 6 additional domains: \emph{alarm}, \emph{messaging}, \emph{music}, \emph{reminder}, \emph{timer}, and \emph{weather}, with a good mixture of simple (flat) and complex (compositional) domains.
Table~\ref{tab:mdtop} shows some basic statistics of the \mdtop{} dataset.
We follow the same process of dataset collection as outlined in the \topdata{} paper.

\begin{table*}
\centering
\small

\begin{tabular} {@{\hspace{0.45em}}r@{\hspace{0.45em}}@{\hspace{0.45em}}l@{\hspace{0.45em}}@{\hspace{0.45em}}c@{\hspace{0.45em}}@{\hspace{0.45em}}c@{\hspace{0.45em}}@{\hspace{0.45em}}c@{\hspace{0.45em}}@{\hspace{0.45em}}c@{\hspace{0.45em}}@{\hspace{0.45em}}c@{\hspace{0.45em}}@{\hspace{0.45em}}c@{\hspace{0.45em}}@{\hspace{0.45em}}c@{\hspace{0.45em}}@{\hspace{0.45em}}c@{\hspace{0.45em}}@{\hspace{0.45em}}c@{\hspace{0.45em}}@{\hspace{0.45em}}c@{\hspace{0.45em}}}
\toprule
 &Target Domain   & \multicolumn{3}{c}{reminder}  && \multicolumn{3}{c}{weather} &&\\
\cmidrule{3-5}\cmidrule{7-9}
\ftextnumero    && \#Train & \#Valid & Accuracy && \#Train & \#Valid & Accuracy && Average \\
\midrule
\multicolumn{9}{l}{\emph{Supervised models with 1000 SPIS}} \\
\rownumber{tabrow:1000spis:lstm} & LSTM-\copyptr{} & 7552 & 2526 &  71.7  && 4197  & 2667 & 81.0  && 76.4 \\[1ex]

\multicolumn{9}{l}{\emph{Supervised models with 500 SPIS}} \\
\rownumber{tabrow:500spis:lstm} & LSTM-\copyptr{} & 4788 & 2526 & 65.9   && 2372 &  2667 &78.6   && 72.3 \\
\rownumber{tabrow:400spis:roberta} & RoBERTa-\copyptr{}~\cite{rongali-etal-2020-dont} & 4788 & 2526 &  71.9  && 2372  & 2667 & 83.5  && 77.7 \\
\rownumber{tabrow:400spis:bart} & BART-\copyptr{} & 4788 & 2526 &  71.9  && 2372  & 2667 & 84.9  && 78.3 \\
\midrule

\multicolumn{9}{l}{\emph{Low-Resource models with 25 SPIS}} \\
\rownumber{tabrow:25spis:lstm:ftonly} & LSTM-\copyptr{} (FT only) & 493 & 337  & 21.5      &&  176 & 147 & 46.2      && 33.8      \\
\rownumber{tabrow:25spis:bart:ftonly} & BART-\copyptr{} (FT only) & 493 & 337   &  55.7     &&  176 & 147  & 71.6      && 63.6    \\[1ex]

\rownumber{tabrow:25spis:bart:jt} & BART-\copyptr{} (JT) & 493  & 337  &  57.1     &&  176  & 147 & 71.0      && 64.1      \\
\rownumber{tabrow:25spis:bart:jt10x} & BART-\copyptr{} (JT 10x) & 493  & 337  & 59.2      &&  176 & 147  & 73.3      && 66.2      \\
\rownumber{tabrow:25spis:bart:jt100x} & BART-\copyptr{} (JT 100x) & 493  & 337  & 58.9      &&  176 & 147  & 74.7      && 66.8      \\[1ex]

\rownumber{tabrow:25spis:lstm:stft} & LSTM-\copyptr{} (ST+FT) & 493  & 337  & 45.8      &&  176 & 147  & 65.1      && 55.4      \\
\rownumber{tabrow:25spis:roberta:stft} & RoBERTa-\copyptr{} (ST+FT) & 493  & 337  & 63.7      &&  176 & 147  & 76.0      && 69.9     \\
\rownumber{tabrow:25spis:bart:stft} & BART-\copyptr{} (ST+FT) & 493  & 337  & 68.0      &&  176 & 147  & 75.9      && 72.0       \\[1ex]

\rownumber{tabrow:25spis:bart:reptile} & BART-\copyptr{} (Reptile+FT) & 493  & 337  & \bf 70.5      &&  176 & 147  & \bf 77.7      && \bf 74.1      \\

\bottomrule
\end{tabular}

\caption{Results on the \mdtop{} dataset. Accuracy: Exact Match Accuracy; ST: Source Training; FT: Fine-Tuning (Target Training); JT: Joint Training; 10x: 10x Target Domain Upsampling. See more details in \secref{sec:exp:result}.}
\label{tab:mdtop-results}
\end{table*}

\subsection{Experimental Setup}\label{sec:exp:setup}

To evaluate our model on low-resource domain scaling for both complex and simple (flat) domains, we use \emph{reminder} and \emph{weather} as the target domains.
The remaining 6 domains are used as source domains.
As mentioned in \secref{sec:model:setup}, to study how much data is needed to achieve a good performance on the target domain, we adopt the SPIS strategy (samples per intent and slot) instead of selecting a fixed number of training samples for each target domain.
In particular, we focus on a low-resource setting of 25 SPIS (see \secref{sec:exp:spis-curve}), where samples are randomly selected to ensure each intent and slot appears in at least 25 training instances.
On the other hand, supervised models are trained with 500 or 1000 SPIS to assess the performance of our low-resource domain scaling model.
For the source domains, all available training data is utilized.

\paragraph{Validation Set}
To perform model selection and early stopping, a validation set is adopted, which is also set to 25 SPIS for simplicity.
In contrast, the supervised models utilize the entire validation set as shown in Table~\ref{tab:mdtop-results}.

\paragraph{Data Preprocessing}
We first perform standard preprocessing such as lower-casing and tokenization.
For models initialized with pre-trained language representations, BPE~\cite{sennrich-etal-2016-neural} tokenization is done to match that used by the pre-trained model.
We do not tokenize ontology tokens (intents and slot labels) into BPE, but instead treat them as atomic tokens which are appended to the BPE vocabulary.

We then perform two additional preprocessing (canonicalization) steps, consistent across all models.
First of all, note that certain utterance tokens do not contribute to the semantics of the query.
For instance, in Figure~\ref{fig:nested-query}, the phrase \emph{Driving directions to} under \texttt{IN:GET\_DIRECTIONS} and \emph{the} under \texttt{IN:GET\_EVENT} can be omitted as their semantics are already captured by the intent labels.
Therefore, we only retain utterance tokens under \emph{leaf slots} (\emph{Eagles} and \emph{game} in Figure~\ref{fig:nested-query}) while removing all others.
Furthermore, we sort the children of each node in the semantic parse tree in alphabetical order of the label, since the order of the children does not alter its semantic meaning.
In the case of Figure~\ref{fig:nested-query}, \texttt{SL:NAME\_EVENT} and \texttt{SL:CAT\_EVENT} will be reordered.
We call the final semantic parse after these two preprocessing steps the \textbf{canonical form}, which will be used in all experiments.

\subsection{Results and Discussions}\label{sec:exp:result}
Our main experimental results are summarized in Table~\ref{tab:mdtop-results}.
LSTM-\copyptr{} utilizes BiLSTMs as both the encoder and decoder, which are commonly adopted in practice due to their smaller model size, faster inference time, and sometimes better performance when training data is sufficient~\cite{rongali-etal-2020-dont}.
RoBERTa-\copyptr{} is the most similar to the model proposed by \newcite{rongali-etal-2020-dont}\footnote{Our implementation (\secref{sec:model:arch}) is not identical to \newcite{rongali-etal-2020-dont}; please refer to their paper for the differences.}, which uses the RoBERTa encoder and a randomly initialized transformer decoder.
BART-\copyptr{} is our proposed model (\secref{sec:model:arch}) which leverages BART to initialize both the encoder and decoder.

On the other hand, FT in the table refers fine-tuning or target training, and a FT only model trains solely on the 25 SPIS training data on a target domain.
In contrast, ST+FT models first go through source training in which the models are trained on all source domain data, and are then fine-tuned on the target domain.
JT stands for joint training, where the training data of the source and target domains are concatenated to jointly train the model.
Since the target domains have very few samples (25 SPIS) compared to the source domains, upsampling can be conducted.
For instance, JT 100x indicates that the target domain samples are duplicated 100 times before concatenated with the source domains.
Finally, Reptile+FT is our meta-learning approach, where standard source training is replaced with Reptile, as described in \secref{sec:metalearning}.

\paragraph{Pre-trained language representations}
As demonstrated in Table~\ref{tab:mdtop-results}, pre-trained representations is crucial in the low-resource setting with a very small amount of training data, where the knowledge encoded in these representations can dramatically improve the model's generalization.
In our experiments, BART outperforms LSTM by 17\% in the ST+FT setting (row \ref{tabrow:25spis:lstm:stft} \& \ref{tabrow:25spis:bart:stft}), and 30\% in the FT only setting (row \ref{tabrow:25spis:lstm:ftonly} \& \ref{tabrow:25spis:bart:ftonly}).

Furthermore, we demonstrate that BART is a superior choice over RoBERTa to initialize our \decoupled{} model, indicating that pretraining both the encoder and decoder works well for semantic parsing, even when the pretraining was based on denoising English sentences without using any logical forms.
Comparing row \ref{tabrow:25spis:roberta:stft} and \ref{tabrow:25spis:bart:stft}, we observe a performance gap of 4.3\% between BART and RoBERTa on the \emph{reminder} domain, which is more complex with more labels and deep compositional structures.
A closer look on the reminder domain also reveals that BART outperforms RoBERTa by a larger margin on compositional queries than flat ones (8.7\% relative improvement on compositional queries vs.~6.3\% on flat; numbers not shown in table).

\paragraph{Importance of source training}
As shown in Table~\ref{tab:mdtop-results}, source training also plays a critical role in low-resource domain scaling.
With the non-pretrained LSTM model, source training can improve the performance from 33.8\% to 55.4\% (row \ref{tabrow:25spis:lstm:ftonly} \& \ref{tabrow:25spis:lstm:stft}), showing that source training can teach the model important inductive biases for the semantic parsing task.
With BART, one hypothesis might be that source training is no longer important as BART learns a sufficiently good representation to provide model generalization. 
Nevertheless, this is not the case as revealed by row \ref{tabrow:25spis:bart:ftonly} and \ref{tabrow:25spis:bart:stft}, where source training improves the BART model's performance by 8.4\%.

One possible explanation is that the model can learn useful knowledge about the semantic spaces (tree structures) as well as certain intents and slots during source training.
For instance, the improved accuracy on \emph{reminder} may be explained in part by its similarity to various source domains such as \emph{alarm} and \emph{timer}.
In addition, the target domains share some common slots with the source domains, such as \texttt{SL:DATE\_TIME} and \texttt{SL:LOCATION}.
When exposed to many more instances of these slot values on the source domains, the model can learn to better capture the semantics of the slots, leading to enhanced performance.

\paragraph{Joint training vs.~fine-tuning}
In this paper, we adopt the source training + fine-tuning strategy.
An alternative is joint training where the training data is combined from the source and target domains.
Optionally, we can also upsample the target domains training data to increase model exposure.
As shown in Table~\ref{tab:mdtop-results} (row \ref{tabrow:25spis:bart:jt}-\ref{tabrow:25spis:bart:jt100x}, \ref{tabrow:25spis:bart:stft}), however, joint training performs worse than ST+FT.
It is a consistent empirical observation yet a curious one that fine-tuning achieves superior performance than joint training, which may deserve further investigation.
Nonetheless, joint training does not suffer from the \emph{forgetting} issue on the source domains, and may be the preferred choice for building a single model for both the source and target domains.

\paragraph{Meta-learning}
Finally, we show that meta-learning can improve model training for transferring to low-resource domains (row \ref{tabrow:25spis:bart:stft} \& \ref{tabrow:25spis:bart:reptile}).
When standard source training is replaced with Reptile, the accuracy of the BART model is substantially improved on both target domains and the best performance is achieved across all low-resource models (+2.1\%).
Even compared to supervised models trained with 500 SPIS, a 10-fold data increase, our Reptile+BART model outperforms the LSTM-based model, and is only 3.6\% away from the state-of-the-art RoBERTa-based one.

\begin{figure}
    \centering
    \includegraphics[width=\linewidth]{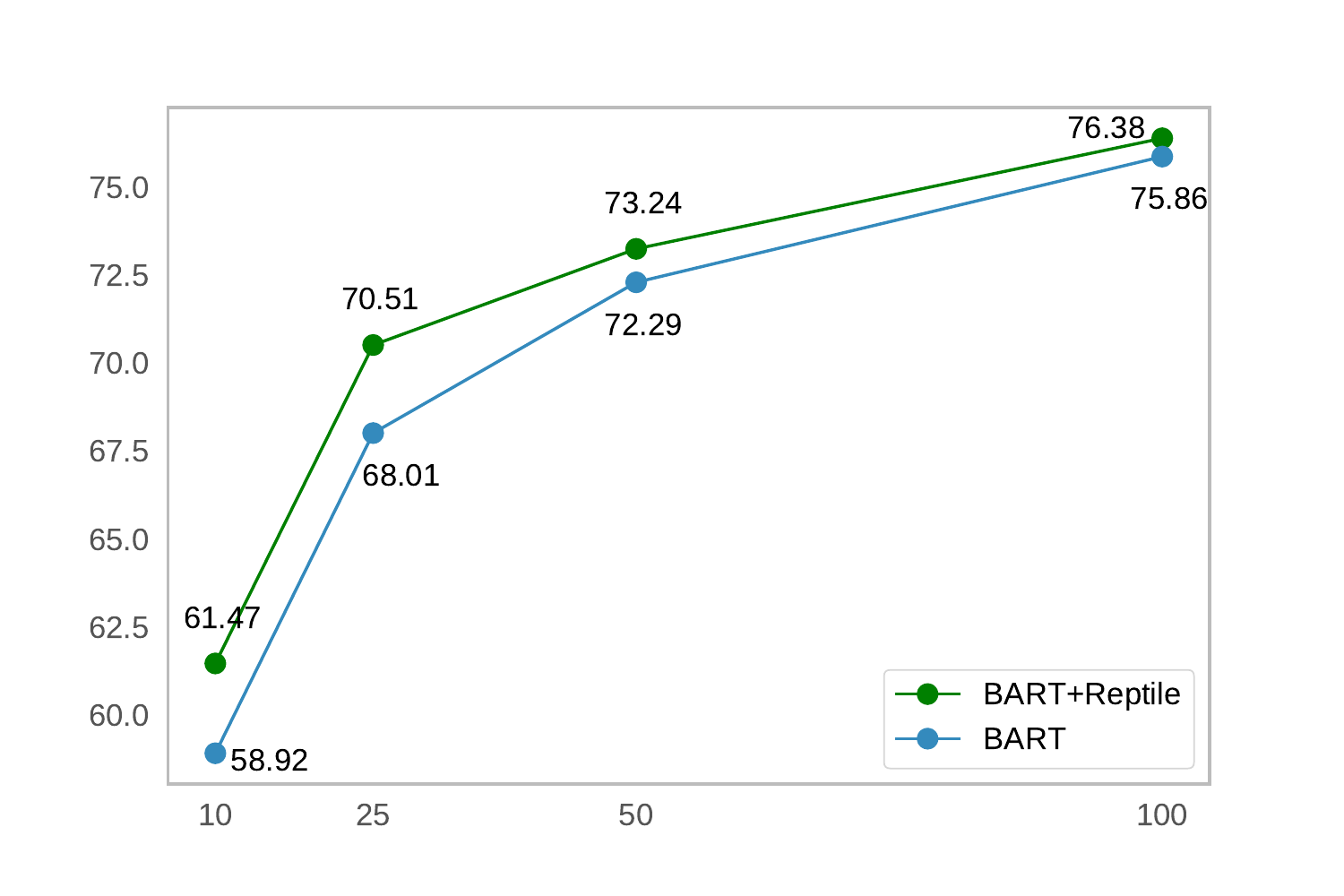}
    \caption{Performance of BART-\copyptr{} (ST+FT) and BART-\copyptr{} (Reptile+FT) at 10, 25, 50 and 100 SPIS on the \emph{reminder} domain.}
    \label{fig:spis-curve}
\end{figure}

\subsection{Accuracy vs.~SPIS}\label{sec:exp:spis-curve}
Figure~\ref{fig:spis-curve} shows the performance on the \emph{reminder} domain of our two best models, BART-\copyptr{} (ST+FT) and BART-\copyptr{} (Reptile+FT), at 10, 25, 50 and 100 SPIS.
For each SPIS setting, we take the model with the best validation set accuracy over 5 runs.
Similar to the main experiment, the validation set is selected using the same SPIS setting as the training set.

First, we observe that meta-learning proves to be beneficial for the extremely low resource setting and improves the performance of BART by about 2.5\% at both 10 and 25 SPIS.
The performance gap is gradually reduced as the amount of training data increases.
Furthermore, we notice a steeper performance drop for both models when we go below 25 SPIS, which gives us an idea of the amount of annotated data required to achieve an acceptable performance on a new domain.
In future work, we plan to push the boundary further to learn effective models with even less training data.

\subsection{Implementation Details}\label{sec:exp:impl}
For the LSTM models, both the encoder and decoder have 2 layers and a hidden size of 512.
Dropout ($p=0.4$)~\cite{JMLR:v15:srivastava14a} is applied.
Adam~\cite{kingma2014adam} is used for optimization, with a learning rate of $10^{-3}$ for source training and $5\times 10^{-4}$ for low-resource fine-tuning.
For RoBERTa models, the encoder is a 12-layer transformer with an embedding size of 768, while the decoder is a smaller transformer model with 3 layers and an embedding size of 256.
For BART models, both the encoder and decoder follow the size of the pre-trained model with 12 layers and an embedding size of 1024.
For all transformer-based models, Dropout ($p=0.3$) is applied.
Adam is again used for optimization, with a learning rate of $10^{-4}$ for source training and $5\times 10^{-5}$ for fine-tuning.
In addition, the inverse square-root learning rate schedule is employed with a warmup period of 4000 updates for source training, and 2000 for fine-tuning.
For meta-learning, we select $k=5$ and a batch size of 32 for Reptile, with both inner ($\eta$) and outer ($\alpha$) learning rates being $5\times 10^{-5}$.

All models are trained for 100 epochs on the source domains with a batch size of 128 (except Reptile), using early stopping if the validation accuracy does not improve in the last 10 epochs.
Fine-tuning is done for 2000 epochs, with a batch size of either 64 (LSTM and RoBERTa) or 32 (BART and meta-learning).
Model validation is performed once every 10 epochs during fine-tuning, and stops early after 20 consecutive validations with no improvements.
Our model is implemented with the \emph{fairseq} framework~\cite{ott-etal-2019-fairseq} and trained on a Nvidia Telsa P100 GPU with 16GB memory.

\section{Related Work}\label{sec:relatedwork}

\noindent\textbf{Task-Oriented Semantic Parsing}
has attracted attention from the research community since 1990s with the advent of the ATIS dataset~\cite{price-1990-evaluation}.
Traditionally, the task is formulated as a joint text classification (intent prediction) and sequence tagging (slot filling) problem, and can be tackled with sequence labeling models such as RNNs~\cite{mesnil2013investigation,Liu+2016}.
These models can only parse \emph{flat} queries with one intent and non-nested slots.
More recently, a number of studies propose alternative approaches for handling the more complex \emph{compositional} queries using neural shift-reduce parsers~\cite{gupta-etal-2018-semantic-parsing,einolghozati2019improving} or seq2seq models~\cite{jia-liang-2016-data,rongali-etal-2020-dont}.

On the other hand, there have been research efforts on scaling task-oriented parsers to new domains with less training data~\cite{Jaech2016DomainAO,Bapna2017TowardsZF,fan-etal-2017-transfer,goyal-etal-2018-fast,DBLP:conf/aaai/LeeJ19}.
These methods, however, only focus on the simpler flat queries.
Our proposed method, in contrast, can effectively parse both flat and compositional queries for low-resource target domains.

\noindent\textbf{Meta-Learning}~\cite{Lake1332}, or learning to learn, aims to learn a model that can quickly adapt to new tasks with a small amount of training data.
In particular, \citet{finn-etal-2017-maml} propose MAML, an optimization-based meta-learning method, which learns a good parameter initialization suitable for faster adaptation to new tasks.
As MAML requires to compute second derivatives, which are computation and memory intensive, there have been studies to use either first-order approximation such as first-order MAML and Reptile~\cite{DBLP:journals/corr/abs-1803-02999}, or implicit differentiation~\cite{NIPS2019_8306}.
Furthermore, meta-learning has also been applied to a number of NLP tasks lately~\cite{gu-etal-2018-meta,dou-etal-2019-investigating,ijcai2019-437,qian-yu-2019-domain,2019arXiv190905438S}.
\section{Conclusion}\label{sec:conclusion}
In this work, we study the low-resource domain scaling problem for task-oriented semantic parsing.
In particular, we focus on the 25 SPIS setting to investigate whether a model can effectively adapt to new \emph{domains} with a very limited amount of training data.
Our approach distinguishes itself from previous methods on two fronts.
First of all, we argue the encoder-only pre-trained representations used in existing work are not ideal for the seq2seq model employed in task-oriented semantic parsing, and instead propose to use BART, a pre-trained model with an encoder-decoder architecture.
More importantly, we adopt optimization-based meta-learning to improve the model's generalization to new target domains with very few training samples.

Our experiments show that our proposed method significantly outperforms all competing methods and achieves the best performance in the low-resource setting.
Even when compared with supervised models trained with 500 SPIS, a 10-fold data increase, our best performing model remains competitive, and outperforms a state-of-the-art LSTM-based Pointer Generator Network~\cite{rongali-etal-2020-dont}.
Last but not least, we collect the \mdtop{} dataset, a large-scale multi-domain task-oriented semantic parsing dataset with 8 domains and more than $180k$ annotated samples to evaluate our models, which we release to the research community.

\bibliographystyle{acl_natbib}
\bibliography{anthology,lrds}

\end{document}